# Multi-lingual and Multi-cultural Figurative Language Understanding

**Anubha Kabra**[1*], **Emmy Liu**[1*], **Simran Khanuja**[1*], **Alham Fikri Aji**[2],
**Genta Indra Winata**[3], **Samuel Cahyawijaya**[4], **Anuoluwapo Aremu**[5],
**Perez Ogayo**[1], **Graham Neubig**[1]
[1]Carnegie Mellon University   [2]MBZUAI   [3]Bloomberg   [4]HKUST   [5]Masakhane

## Abstract

Figurative language permeates human communication, but at the same time is relatively understudied in NLP. Datasets have been created in English to accelerate progress towards measuring and improving figurative language processing in language models (LMs). However, the use of figurative language is an expression of our cultural and societal experiences, making it difficult for these phrases to be universally applicable. In this work, we create a figurative language inference dataset, MABL, for seven diverse languages associated with a variety of cultures: Hindi, Indonesian, Javanese, Kannada, Sundanese, Swahili and Yoruba. Our dataset reveals that each language relies on cultural and regional concepts for figurative expressions, with the highest overlap between languages originating from the same region. We assess multilingual LMs' abilities to interpret figurative language in zero-shot and few-shot settings. All languages exhibit a significant deficiency compared to English, with variations in performance reflecting the availability of pre-training and fine-tuning data, emphasizing the need for LMs to be exposed to a broader range of linguistic and cultural variation during training.[1]

## 1 Introduction

When you are feeling happy, do you think that you are "warm" or "cold"? If you are a monolingual English speaker, you will likely answer "warm", and use expressions like "this really warmed my heart". However, if you are a native Hindi speaker, you may answer "cold", and use expressions like दिल को ठंडक पढ़ना ("coldness spreads in one's heart" ) (Sharma, 2017). Linguistic communication often involves figurative (i.e., non-literal) language (Shutova, 2011; Fussell and Moss, 2008;

|    | Figurative Expression | Inference |
|----|----------------------|-----------|
| yo | Omah iku kaya istana (The house is like a palace.) | Omah iku apik banget. (The house is very nice.) <br> Omah iku elek banget. (The house is very ugly.) |
| id | Rambutnya seperti bihun. (Her hair is like vermicelli.) | Rambutnya keriting. (Her hair is curly.) <br> Rambutnya lurus. (Her hair is straight.) |
| hi | जीवन मीठा गुलकन्द है। (Life is sweet Gulkand. ) | जीवन अच्छा है। (Life is good.) <br> जीवन बुरा है। (Life is bad.) |
| kn | ಅದು ದೋಸೆಯಂತೆ ಗರಿಗರಿಯಾಗಿತ್ತು. (It was crispy like a dosa.) | ಅದು ಗರಿಗರಿಯಾಗಿದೆ (It is crisp.) <br> ಅದು ಗರಿಗರಿಯಾಗಿರಲಿಲ್ಲ (It was not crisp.) |
| sw | Maneno yake ni sumu. (His words are like poison.) | Maneno yake yanaponya. (His words heal.) <br> Maneno yake yanaangamiza. (His words are devastating.) |

Table 1: Examples of figurative expressions and respective inferences from the collected data. Correct answers are highlighted in green.

Lakoff and Johnson, 1981), which is laden with implicit cultural references and judgements that vary cross-culturally. Differences in figurative expressions used in different languages may be due to cultural values, history, or any number of other factors that vary across where the languages are spoken.[2] Understanding figurative language therefore relies on understanding what concepts or objects are considered culturally significant, as well as their sentiment in that culture.

Better understanding of figurative language would benefit tasks such as hate speech detection or sentiment classification (ElSherief et al., 2021; van Aken et al., 2018). However, state-of-the-art language models have been shown to frequently misinterpret both novel figurative expressions and conventionalized idioms, indicating the need for improved methods (Dankers et al., 2022;

---

* These authors contributed equally.
[1]Data and code is released at https://github.com/simran-khanuja/Multilingual-Fig-QA

[2]The Hindi example is most likely attributable to climatic conditions, as cold may be seen as comparatively more positive in an area where extreme heat is more common (Sharma, 2017)

Liu et al., 2022). Most empirical results probing language models' abilities with respect to figurative language have been based on data in English, meaning there is a comparative lack of resources and study in other languages (Chakrabarty et al., 2022; Liu et al., 2022; Pedinotti et al., 2021a).

We find English figurative language datasets may not have cultural relevance for other languages (§2). This is a general challenge in NLP, as assumptions of common knowledge and important topics to talk about vary from culture to culture (Hershcovich et al., 2022). In order to better train multilingual models to interpret figurative language, as well as to understand linguistic variation in figurative expressions, we construct a multilingual dataset, MABL (**M**etaphors **A**cross **B**orders and **L**anguages), of 6,366 figurative language expressions in seven languages (§3). Examples are shown in Table 1.

We use the dataset to conduct a systematic analysis of figurative language patterns across languages and how well they are captured by current multilingual models (§4). We find that figurative language is often very culturally-specific, and makes reference to important entities within a culture, such as food, mythology, famous people, or plants and animals native to specific regions.

We benchmark multilingual model performance (§5) and analyze model failures (§6), finding that zero-shot performance of multilingual models is relatively poor, especially for lower-resource languages. According to (Liu et al., 2021), main factors which poses challenges on the performance in such cases are cross-lingual transfer and concept shift across languages. However, we observe that concept shift seems to play a larger role due to culturally specific examples. Adding a few examples in the target language can improve performance of larger models, but this is more beneficial for lower-resource languages. This highlights the importance of including culturally relevant training data, particularly data that highlights not just the existence of a concept, but also how people view that concept within that culture.

## 2 Linguistic and Cultural Biases of Existing Figurative Language Datasets

To confirm the importance of building a multilingual, multi-cultural figurative language dataset, we first performed a pilot study to examine the feasibility of instead translating an existing figurative language dataset, Fig-QA (Liu et al., 2022), from

| Lang. | fr | hi | ja |
|---|---|---|---|
| Incorrect | 13% | 40% | 21% |
| Culturally irrelevant | 17% | 20% | 17% |

Table 2: Correctness and cultural relevance of Google translations of Fig-QA validation set.

English into other languages. While there are well-known problems with using translation to create multilingual datasets for tasks such as QA (Clark et al., 2020), it is still worth examining these issues in the context of figurative language in particular. We used the Google Translate Python API to translate the development set into languages that the authors of this paper understood.[3] These were French, Japanese, and Hindi. Each annotator annotated 100 examples for both correctness (whether or not the translation was accurate), and cultural relevance (whether or not the expression was one that would make sense to a native speaker from the culture where the language is predominant).

As seen in Table 2, the number of incorrect examples is large, particularly for Hindi and Japanese. This is mainly due to expressions that don't translate directly (such as a "sharp" conversation in English). Culturally irrelevant examples are due to implicitly assumed knowledge. For instance, a crowdworker from the US generated the example "it's as classic as pancakes for breakfast" with the meaning "it's very classic". However, most people from Japan would not see pancakes as a traditional breakfast, and the meaning "it's not classic" would be more appropriate.

The shift in topics discussed in cultures associated with different languages can be captured by native speakers familiar with that culture, motivating our collection of natural figurative language examples from native speakers.

## 3 The MABL Dataset

### 3.1 Language Selection

We choose the following seven languages: Hindi (hi), Yoruba (yo), Kannada (kn), Sundanese (su), Swahili (sw), Indonesian (id), and Javanese (jv).

The factors we considered while choosing these languages are as follows :

**i)** We aimed to include a range of languages representing the different classes in the resource-based taxonomy of languages, proposed by Joshi et al. (2020), subject to annotator availability.

---
[3]https://pypi.org/project/googletrans/

| Language | #Samples |
|---|---|
| id | 1140 |
| sw | 1090 |
| su | 600 |
| jv | 600 |
| hi | 1000 |
| kn | 1190 |
| yo | 730 |

Table 3: Number of collected samples per language.

**ii)** We chose languages with a sizeable speaker population as shown in Table 5.

**iii)** Our languages come from 5 typologically diverse language families spoken in 4 different countries, which allows us to include a wide range of linguistic and cultural diversity in our data.

Details about the characteristics of each language in terms of available training data and number of speakers can be found in Table 5. Additional information on linguistic properties of these languages can be found in Appendix A.

### 3.2 Dataset Collection

To create culturally relevant examples, we crowd-sourced sample collection to two or more native speakers in the seven languages. The workers were asked to generate paired metaphors that began with the same words, but had different meanings, as well as the literal interpretations of both phrases.

Workers were not discouraged from generating novel metaphors, but with the caveat that any examples should be easily understood by native speakers of that language, e.g., "it's as classic as pancakes for breakfast" would not be valid if pancakes are not a breakfast food in the country in which that language is spoken.

Instructions given to annotators can be found in Appendix B. After collection, each sample was validated by a separate set of workers who were fluent in that language. Any examples that were incoherent, offensive, or did not follow the format were rejected. The number of samples collected per language can be seen in Table 3. Examples of collected data can be seen in Table 1. We note that because of the limited number of samples in each language, we view the samples collected as a *test set* for each language, meaning there is no explicit training set included with this release.

## 4 Dataset Analysis

### 4.1 Concepts expressed

In the structure mapping theory of metaphor, figurative language involves a **source** and **target** concept, and a comparison is made linking some features of the two (Gentner, 1983). Following Liu et al. (2022), we refer to the source as the "subject" and target as "object".[4]

We expect objects referenced to be quite differently cross-culturally. We confirm this by translating sentences from our dataset into English, then parsing to find objects. The number of unique concepts per language, including examples, is listed in Appendix C. This may overestimate the number of unique concepts, as some concepts may be closely related (e.g., "seasonal rain" vs. "rainy season"). Despite this, we are able to identify many culturally specific concepts in these sentences, such as specific foods (hi: samosa, hi: sweet gulkand, id: durian, id: rambutan), religious figures (kn: buddha's smile, sw: king soloman), or references to popular culture (id: shinchan, yo: aníkúlápó movie, en: washington post reporter).

We observe that, excluding pronouns, only 6 objects are present in all languages. These are {"sky", "ant", "ocean", "fire", "sun", "day"}. Of course, variations of all these concepts and other generic concepts may exist, since we only deduplicated objects up to lemmatization, but this small set may indicate that languages tend to vary widely in figurative expressions. Appendix D indicates the Jaccard similarity between objects in each language, which is an intuitive measure of set similarity. The equation is also given below for sets of objects from language A ($L_A$) and langugage B ($L_B$).

$$J(L_A, L_B) = \frac{|L_A \cap L_B|}{|L_A \cup L_B|} \quad (1)$$

The most similar language based on concepts present is highlighted in Table 4. Languages from the same region tend to group together. The set of concepts in English is actually most similar to Swahili.[5] Upon inspection, there were many general terms related to nature, as well as many references to Christianity in the Swahili data, which may explain the similarity to English.[6]

---

[4]This terminology may be confusable with subject and object in linguistics, but was used because the source and target tend to appear in these linguistic positions in a sentence.
[5]There are no particularly closely related languages to English in our dataset
[6]Authors of this paper examined unique concepts expressed in English, Swahili, and Kannada. Swahili sentences had

| Lang. | hi | id | jv | kn | su | sw | yo | en |
|---|---|---|---|---|---|---|---|---|
| Most similar | kn | jv | sw | hi | jv | hi | sw | sw |

Table 4: Most similar concepts sets for each language, based on Jaccard similarity of objects in each language's sentences. Note that as in Appendix A, {hi, kn}, {id, jv, su} and {sw, yo} respectively occur in similar geographic regions.

| Lang. | Speakers (M) | Training Data (in GB) | | Class |
|---|---|---|---|---|
| | | XLM-R | mBERT | |
| en | 400 | 300.8 | 15.7 | 5 |
| hi | 322 | 20.2 | 0.14 | 4 |
| id | 198 | 148.3 | 0.52 | 3 |
| jv | 84 | 0.2 | 0.04 | 1 |
| kn | 44 | 3.3 | 0.07 | 1 |
| su | 34 | 0.1 | 0.02 | 1 |
| sw | 20 | 1.6 | 0.03 | 2 |
| yo | 50 | - | 0.012 | 2 |

Table 5: Per-language statistics (including en for reference); the speaker population of each language, its representation in pre-trained multilingual models, and the Joshi et al. (2020) class each language belongs to. First-language speaker population information is obtained from Wikipedia and Aji et al. (2022). We obtain data size estimates for multilingual BERT from Wikipedia 2019 dump statistics.[7]

### 4.2 Commonsense Categories

We follow the commonsense categories defined in Liu et al. (2022) to categorize knowledge needed to understand each sentence: physical object knowledge (obj), knowledge about visual scenes (vis), social knowledge about how humans generally behave (soc), or more specific cultural knowledge (cul). The same sentence can require multiple types of knowledge. Table 6 shows the prevalence of each type of commonsense knowledge as documented by annotators. Social and object knowledge are the most dominant types required, with Yoruba having an especially high prevalence of social examples. Not many examples were marked as cultural. This may be due to differences in what annotators viewed as cultural knowledge: some knowledge may be considered to fall under the object or social category by annotators, but these same examples may seem culturally specific to people residing in the United States because the objects referenced are not necessarily relevant to English speakers in the US.

---

[†] 18/481 Christianity related concepts, while English had 13/954. Kannada did not have any Christianity related concepts but rather concepts related to Hinduism.

| Lang. | Object | Visual | Social | Cultural |
|---|---|---|---|---|
| hi | 52.4 | 16.4 | 42.0 | 9.2 |
| id | 45.8 | 5.7 | 45.6 | 7.5 |
| jv | 34.0 | 15.0 | 43.3 | 10.0 |
| kn | 63.3 | 17.1 | 20.3 | 15.2 |
| su | 34.3 | 8.6 | 33.3 | 24.0 |
| sw | 48.0 | 20.2 | 32.2 | 5.6 |
| yo | 37.3 | 6.1 | 81.0 | 10.7 |

Table 6: Proportion of common-sense categories.

### 4.3 Cross-lingual concept distribution

To better understand the linguistic and cultural distribution of examples, we extract sentence-level representations from two models: **i)** XLM-R$_{large}$ (Conneau et al., 2019), our best performing baseline model; and **ii)** LaBSE (Feng et al., 2020), a language-agnostic sentence embedding model, optimized for cross-lingual retrieval. We observed that XLM-R clusters by language, whereas LaBSE clusters sentences from multiple languages together, based on conceptual similarity (as shown in Figure 2). Since LaBSE is optimized for cross-lingual sentence similarity, we chose the latter to conduct further analysis.

First, we probe different edges of the cluster and observe concepts along each edge, as visualized in Figure 1. For each concept, we observe sentences from various languages clustering together. Further, these sentences portray cultural traits pertaining to each language. For example, *rice* is commonly mentioned in languages from Indonesia, given that it is a staple food product there.[8] Other examples include sentences in Hindi such as *This house is as old as a diamond* (*diamonds* have a significant historical background in India) or *Your house is worth lakhs* (*lakh* is an Indian English term).[9]

To qualitatively study cultural references, we further analyse metaphors belonging to universal concepts such as *food*, *weather/season*, and *friendship*, searching for sentences containing these keywords.[10] We obtain 230 sentences containing *food*, 111 sentences containing *weather/season* and 307 sentences containing *friend*. A few examples are as shown in Table 7. We observe multiple regional and cultural references, which may not be under-standable by non-native speakers. For example, annotators make references to the *weather/season* with *Peacock* and *frying fish on asphalt* which are innate comparisons in su. With reference to *food*, Indian food commonly uses *Neem* and *Tamarind* as referenced by metaphors in kn and hi. *Neem* is a bitter medicinal herb and *Tamarind* is used to add sourness to food. Finally, we see references to mythological and fictional characters across *friendship* metaphors, where annotators draw from their attributes to describe friendships.

---

[8] https://www.indonesia-investments.com/business/commodities/rice/item183
[9] https://en.wikipedia.org/wiki/Indian_numbering_system
[10] We do a regex search over the word and its translation in each language to obtain sentences from all languages in the concept, using https://projector.tensorflow.org/

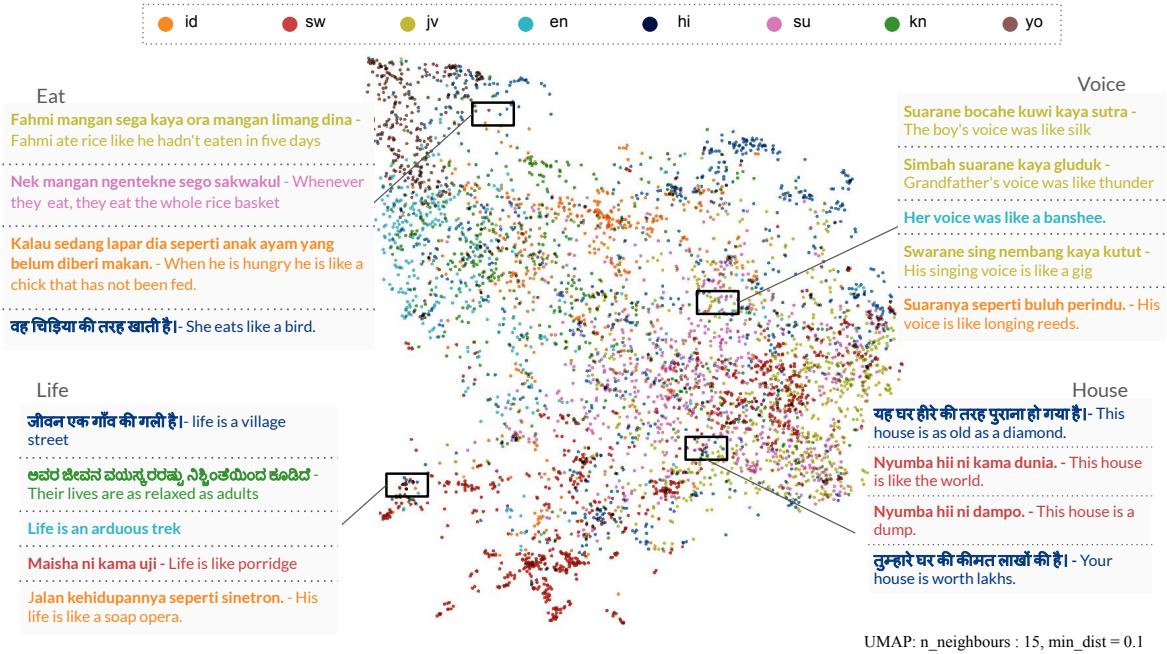

Figure 1: UMAP visualization of the collected data. Sentence embeddings are obtained using LaBSE (Feng et al., 2020), a multilingual dual encoder model, optimized for cross-lingual retrieval. Refer to Section 4 for more details.

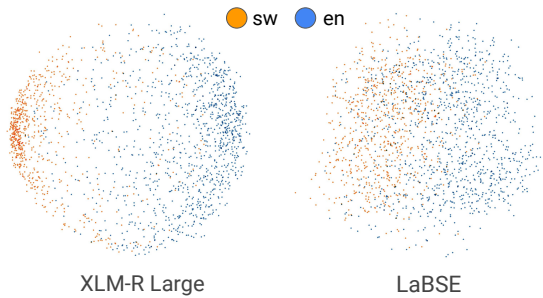

Figure 2: We visualize sentence embeddings for two languages, Swahili (sw) and English (en), using our best-performing model, XLM-R Large (left) and LaBSE (right). Given that en shares the highest number of concepts with sw, we'd expect a tight integration of embedding spaces, which is better displayed by LaBSE.

## 5 Evaluation and Results

### 5.1 Zero-shot

#### 5.1.1 Zero-shot evaluation

Here, we simply fine-tune the Multilingual Pre-trained Language Models (MPLMs) on the English labelled data and evaluate on all target languages. This was performed in the standard format of inputting each example as [CLS] [sentence] [SEP] [meaning1] [SEP] [meaning2] and using a linear layer on the [CLS] token to classify the answer.

#### 5.1.2 Zero-shot transfer results

We present zero-shot evaluation results in Table 8, noting that there can be two contributors to the gap in performance in these seven languages as compared to English. First, since our fine-tuning language is English, there can be a drop in performance simply due to cross-lingual transfer. Second, there is a concept shift in these metaphors, as evidenced by our analysis in Section 4. To discern the contribution of both, we machine-translate the target test sets to en (we refer to this as translate-test). The difference between translate-test and zero-shot, can be thought of as the cross-lingual transfer gap, while the rest of the difference between translate-test and en test performance can be attributed to the concept shift. Due to possible MT errors, the results here represent upper bounds for concept shift and cross-lingual shift, which is further discussed in Section 6.1.

| | References to weather/season | | References to food | | References to friendship |
|---|---|---|---|---|---|
| su | The Indian Ocean is sparkling like a Peacock this Christmas season. | kn | That food is as sweet as Neem. | jv | My friend's father is like a raden werkudara. |
| kn | The weather is also warm like the rainy season. | hi | Hotel food was like tamarind. | hi | He guided his friend like Krishna. |
| su | The weather looks like you can fry fish on the asphalt. | sw | His waist is the width of a baobab. | sw | His friend is abunuwasi. |
| hi | Tina and Ravi's love is like monsoon season. | jv | The taste of this food is like boiled tempeh. | id | He asks the help of his friends just like the king of Tanah Djawo Kingdom. |

Table 7: Translated examples with cultural references specific to regions where these languages are spoken.

**The concept shift gap is generally greater than the cross-lingual gap.** As reported in Table 8, the concept shift gap is greater than the cross-lingual transfer gap for all languages except Swahili, across all models. This result for sw corroborates our findings in Section 4, where we observe that en shares the greatest proportion of object concepts with sw. Given Swahili's extremely low-representation in MPLMs (Table 5), and its high concept overlap with English, we cover most of the gap by simply translating sw to en. For Indonesian (id), we observe that zero-shot performance itself is close to en performance (83.6%) for XLM-R, since id is well-represented in this model (Table 5). Hence, translating to en does not help, and the model needs to be competent in better understanding the cultural references specific to id. In mBERT however, id is poorly represented, and translating to en does help improve performance.

**Performance increases as model and training data size increase, but moreso for higher resource languages.** The smallest model examined, mBERT, has relatively poor performance for all languages, as all languages have < 60% accuracy. Hindi and Indonesian, the two highest-resource languages in our dataset, show a high gain in performance when using a larger model, increasing to 67.58% and 78.09% accuracy respectively. This is especially true for Indonesian, which has a relatively high amount of training data as shown in Table 5. However, lower resource languages tend to show a more modest gain in performance.

## 5.2 Few-shot

### 5.2.1 Few-shot evaluation

While it is common to fine-tune MPLMs on English, given its widespread use and availability, several past works have shown how this is suboptimal (Lin et al., 2019; Debnath et al., 2021) and choosing optimal transfer languages is an important research question in itself (Dhamecha et al., 2021). While the design of an ideal allocation of annotation resources is still unknown, Lauscher et al. (2020) demonstrate the effectiveness of investing in few-shot (5-10) in-language task-specific examples, which provides vast improvements over the zero-shot setup.

We include between 2-50 labelled pairs of sentences from each target language, in addition to the English labelled data, for fine-tuning the model. Training details for all models can be found in Appendix E.

### 5.2.2 Few-shot results

Figure 3 presents the effects of few-shot transfer for each language. Generally, the performance gain is modest. This aligns with results from Lauscher et al. (2020), who found that performance gains were quite small on XNLI. As our task is also an NLI task, we may expect similar improvements. However, we find collecting some cultural examples could disproportionately help low-resource languages.

**Augmenting with few examples usually does not help much** We observed that with a few exceptions, the increase in accuracy on the test set gained was small (< 1%). This is likely because of the diversity of facts needed in order to improve performance. As noted in Section 4.1 and Table 1, this dataset contains many unique cultural references that do not repeat, limiting the utility of seeing a few examples.

**Lower-resource languages benefit more greatly from augmentation** However, there are a few exceptions to this trend. In particular, adding 50 paired Kannada examples to XLM-R$_{large}$ improved performance by 3.83%. Swahili also improves by 1.10% with 50 additional examples for XLM-R$_{base}$, and Sundanese improves by 2.33% with 50 examples for mBERT$_{base}$.

### 5.3 Evaluation of Large Language Models

In addition to the three MPLMs we examine in detail, we also examine the zero-shot performance of large pretrained language models. We choose to examine GPT-3 (text-davinci-003) and BLOOM-176B. As these models are autoregressive rather than masked models, we follow the standard procedure of prediction via choosing the answer with a higher predicted probability (Jiang et al., 2021).

| Model | Language | Zero-shot Performance | Translate-test (to EN) | Cross-Lingual Transfer Gap | Concept Shift Gap |
|---|---|---|---|---|---|
| XLM-R$_{large}$ | en$_{dev}$ | 81.50 ±2.41 | 81.50 ±2.41 | 0.00 | 0.00 |
| | hi | 67.58 ±1.38 | 67.82 ±1.52 | 0.24 | **13.68** |
| | id | 78.09 ±1.14 | 77.51 ±0.91 | -0.58 | **3.99** |
| | jv | 60.93 ±1.95 | 68.13 ±1.66 | 7.20 | **13.37** |
| | kn | 58.08 ±2.10 | 63.67 ±0.98 | 5.59 | **17.83** |
| | su | 60.40 ±1.98 | 70.07 ±0.92 | 9.67 | **11.43** |
| | sw | 58.16 ±0.73 | 75.29 ±2.05 | **17.13** | 6.21 |
| | yo | - | - | - | - |
| XLM-R$_{base}$ | en$_{dev}$ | 75.26 ±0.95 | 75.26 ±0.95 | 0.00 | 0.00 |
| | hi | 62.48 ±0.31 | 63.29 ±0.84 | 0.81 | **11.97** |
| | id | 68.88 ±0.71 | 66.54 ±1.22 | -2.34 | **9.26** |
| | jv | 53.67 ±0.54 | 58.17 ±0.82 | 4.50 | **17.09** |
| | kn | 54.67 ±1.31 | 57.86 ±1.10 | 3.20 | **17.40** |
| | su | 52.41 ±1.79 | 61.33 ±0.68 | 8.93 | **13.93** |
| | sw | 52.73 ±1.38 | 65.77 ±1.82 | **13.04** | 7.31 |
| | yo | - | - | - | - |
| mBERT$_{base}$ | en$_{dev}$ | 70.88 ±2.46 | 70.88 ±2.46 | 0.00 | 0.00 |
| | hi | 51.32 ±0.94 | 59.45 ±1.77 | 8.13 | **11.43** |
| | id | 56.56 ±1.66 | 63.30 ±1.12 | 6.74 | **7.58** |
| | jv | 55.06 ±1.70 | 60.76 ±2.31 | 5.70 | **10.12** |
| | kn | 52.63 ±1.15 | 56.70 ±0.77 | 4.07 | **14.18** |
| | su | 52.87 ±1.67 | 59.37 ±2.37 | 6.51 | **11.51** |
| | sw | 52.12 ±1.09 | 63.57 ±0.78 | **11.45** | 7.31 |
| | yo | 50.52 ±1.04 | 50.60 ±1.28 | 0.08 | **20.28** |
| text-davinci-003 | en$_{dev}$ | 74.86 | 74.86 | 0.00 | 0.00 |
| | hi | 50.60 | 59.62 | 9.02 | **15.24** |
| | id | 64.21 | 66.93 | 2.72 | **7.93** |
| | jv | 51.00 | 62.17 | 11.17 | **12.70** |
| | kn | 50.08 | 57.85 | 7.76 | **17.02** |
| | su | 49.67 | 58.33 | 8.67 | **16.53** |
| | sw | 54.83 | 65.33 | **10.51** | 9.53 |
| | yo | 50.27 | 48.77 | -1.51 | **26.10** |

Table 8: Averaged zero-shot evaluation ± standard deviation of MPLMs (and GPT-3) across five seeds on all seven languages: Hindi (hi), Indonesian (id), Yoruba (yo), Kannada (kn), Sundanese (su), Swahili (sw), Javanese (jv). Additionally, we translate each of these test sets to EN (translate-test). This helps discern the gap in performance due to *i) cross-lingual transfer* and *ii) concept shift in metaphors.*. These gaps are calculated using the EN validation set's performance as a gold reference. Refer to Section 5.1 for more details. The gap that is higher (which indicates a more significant challenge) is highlighted for each model and language. Note that results for Yoruba are not reported for XLM-R, as it was not trained on any Yoruba data.

The performance of GPT-3 is not very good on most languages when tested zero-shot, but we note that it has a reasonable zero-shot performance on the English development set (74.86%), higher than the reported results of text-davinci-002. (Liu et al., 2022). There is a high concept shift gap as with the other models but also a comparatively higher cross-lingual gap as this model is much stronger in English.

## 6 Error Analysis

### 6.1 Effect of English MT

As noted in Section 5.1, there are two major factors that can cause difficulty in cross-lingual transfer: language shift and concept shift. We try to approximate these effects by translating the test set in each language to English. However, this is done with machine translation, so there may be errors. Despite this, translation can still benefit the model if the original language was low-resource. We can divide the model performance into four cases as shown in Table 9.

| | | Translate-EN | |
|---|---|---|---|
| | | Correct | Incorrect |
| Orig. | Correct | 53.06% | 15.52% |
| | Incorrect | 19.09% | 12.33% |

Table 9: Confusion matrix of examples that were answered correctly by XLM-R$_{large}$ before and after translation to English, across all languages combined.

First, there are easy examples (53%) which are answered correctly in both the original language and translated versions. Next there are linguisti-cally challenging examples (19%) which are originally answered incorrectly, but switch to being answered correctly after being translated to English.[11] There are difficult-to-translate or incorrectly translated examples (15%). It's likely that these errors can be completely eliminated with a careful enough translation. Lastly, there are hard examples (12%) which are answered incorrectly before and after being translated. These contain many inherently difficult examples, and examples with specific cultural terms. Examples of each type can be found in Appendix G.

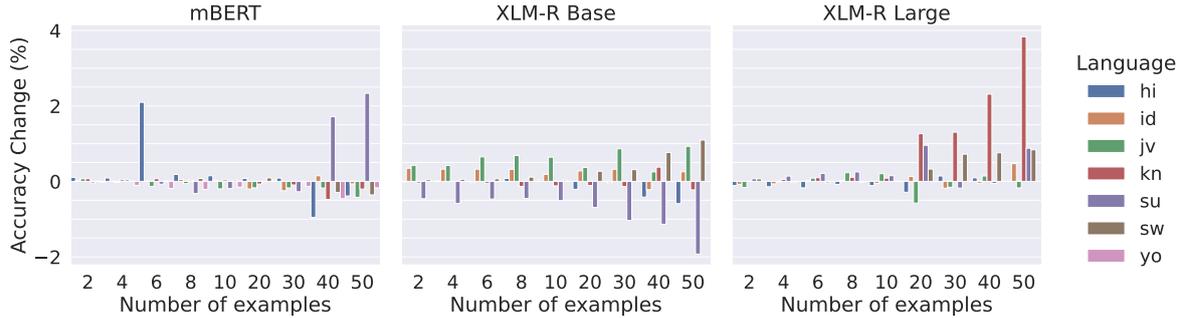

Figure 3: Effect of adding up to 50 examples in the target language to the English training data. This strategy is most beneficial for XLM-R$_{\text{large}}$ with more than 10 examples in the target language. Exact results can be found in Appendix F.

### 6.2 Cultural Examples

We examine the accuracy of XLM-R$_{\text{large}}$ on the commonsense categories in Section 4.2. Overall, there is a small difference in accuracy between cultural examples and the overall accuracy, with overall accuracy at 63.99% and accuracy on cultural examples at 61.68%. Accuracy for all languages can be found in Appendix H. This is a preliminary analysis, but may indicate that references to explicit named entities may not be the only issue for the model with regard to culture.

## 7 Related Work

### 7.1 Figurative Language

**English-centric**: Most previous inference tasks on figurative language have been in English (Chakrabarty et al., 2022; Liu et al., 2022; Pedinotti et al., 2021a). Further, research on figurative language in English centers around training models to detect the presence of metaphors in text (Leong et al., 2020; Stowe and Palmer, 2018;

---
[11] Linguistically challenging here means that the language is more challenging for an LM to perform well in, not that the linguistic structure is very difficult.

Tsvetkov et al., 2014). This is done using datasets primarily consisting of idioms and conventionalized metaphors. However, recognizing common metaphorical phrases may not truly test a model's ability to interpret figurative language. There is limited research on understanding metaphors, which mostly looks at linking metaphorical phrases to their literal meanings through paraphrase detection (Bizzoni and Lappin, 2018) or generation (Shutova, 2010; Mao et al., 2018). Some studies investigate LMs' ability to understand metaphors, but they do not consider the fact that metaphors have different meanings based on context (Pedinotti et al., 2021b; Aghazadeh et al., 2022). Most recently, Liu et al. (2022) released a dataset which requires a model to infer the correct meaning of metaphor, rather than simply identifying or paraphrasing it, hence calling to test deeper semantic understanding.

**Extension to Multilingual**: Research in corpus linguistics (Díaz-Vera and Caballero, 2013; Kövecses, 2004; Charteris-Black and Ennis, 2001) suggests that there significant variation in metaphorical language between cultures. There has been some work in detecting metaphors in multilingual text (Tsvetkov et al., 2013; Shutova et al., 2017). These works have focused on three relatively high-resource languages: English, Russian and Spanish. Both focused on cross-lingual techniques to identify metaphors from newspapers and dictionaries. Hence, there hasn't been any large-scale multilingual dataset of figurative language constructed, which would allow one to study cultural variations across metaphors. We fill this gap with the release of our dataset.

# 8 Conclusion

Despite being relatively widespread, figurative language is relatively under-studied in NLP. This is especially true for non-English languages. To enable progress on figurative language processing, we create MABL, a figurative inference dataset across seven languages. We find considerable variation in figurative language use across languages, particularly in the unique objects that people invoke in their comparisons, spanning differences in food, mythology and religion, and famous figures or events. This variation is likely due to differences in cultural common-ground between the countries in which these languages are spoken. We find that multilingual models have considerable room for improvement on this task, and cross-cultural shift may play a significant role in the performance degradation from English. We encourage the NLP community to further examine the role that culture plays in language, and note that figurative language can be used as a testbed to examine cross-linguistic and cross-cultural variations.

# 9 Limitations

First, despite our pursuit of attempting to understand figurative language use across cultures, we have barely scratched the surface in terms of diverse representation. Due to limited scope, budget, and resources, we collect data from 2-3 annotators per language, for seven languages. Further, culture can vary greatly within a language (Hershcovich et al., 2022). Therefore, until we can represent all of the worlds' people and their languages, there will always be room for improvement.

We also acknowledge that the syntax captured in the dataset may not be the most diverse, as many examples follow the template "<X> is like <Y>". However, we create these simpler examples as a first step, since extension to more complex and naturalistic language can be included in future work.

Second, to analyse concept shift, we machine translate test sets into English. However, these translations can be erroneous to varying degrees, which may have resulted in an over-estimation of error attribution to concept shift. This could not be avoided however, due to limited resources of obtaining human translations.

Third, English may not be the best language to transfer from in zero-shot evaluation of multilingual models. While we were constrained by training data availability, past works have shown that machine-translating train sets can help, an avenue we haven't explored here. Even though we experiment with few-shot evaluation, there may exist an optimal combination of source languages which best transfer to our target languages.

Fourth, the English authors recognized culture-specific terms that were not marked as cultural by annotators in the commonsense categorization across all languages. This may be because annotators, being mostly familiar with their own cultures, attributed culturally specific facts and terms as being common sense. Likewise, the English-speaking participants may have viewed a separate set of facts as common sense which would not be agreed upon by people from a different culture. It is thus difficult to disentangle common sense and culture in many cases.

## A  Selected Languages

Table 10 contains additional information on languages included in the dataset. Information on languages was collected from the World Atlas of Language Structures (WALS) and Glottolog 4.7 (Hammarström et al., 2022; Dryer and Haspelmath, 2013).

## B  Instructions for Annotators

In Liu et al. (2022), workers are prompted with random words taken from English metaphorical frames in Lakoff and Johnson (1981). However, as these metaphorical frames are not readily available in other languages, and we did not want to bias workers toward concepts that are only relevant in English, we chose to omit this prompt and have workers generate sentences freely, while encouraging them to emphasize aspects of their culture. Annotators were paid according to their proposed hourly range ($25/hour on average, all above $15/hr). Validators were paid $15/hr. This study was approved by our IRB. No identifying information was collected.

Note that this is the English version of the instructions, as instructions were machine-translated to each target language and corrected by native speakers.

> Your task is to generate pairs of sentences with opposite or very different meanings, both of which contain metaphors. You can feel free to incorporate creativity into the metaphors, but also make sure that they're something that could be understood by the speakers of the language that you are generating metaphors for, e.g., "this is as classic as pancakes for breakfast" to mean "this is classic" wouldn't make sense for a culture in which pancakes aren't traditionally eaten for breakfast.
>
> You can do this by thinking of a metaphor that conveys a certain meaning, and replacing the metaphorical phrase with another metaphorical phrase of the same type (for instance, noun phrases, verb phrases or adjective phrases) that conveys the opposite meaning.
>
> Here are some examples of metaphors to give you an idea of what we're looking for: Please write both the metaphor and its meaning for each sentence.
>
> 1. The surgeon is (a lumberjack/a ballet dancer).
> 2. The movie has the depth of (a wading pool/the grand canyon)
> 3. Her commitment to the cause was as sturdy as (plywood/oak)
>
> If you're stuck, a general template you can use is <SUBJECT> is <metaphor 1>/<metaphor 2>.

| Language | Branch | Countries | Word Order |
|---|---|---|---|
| Hindi | Indo-European | India | SOV |
| Indonesian | Austronesian | Indonesia | SVO |
| Javanese | Austronesia | Indonesia | SVO |
| Kannada | Dravidian | India | SOV |
| Sundanese | Austronesian | Indonesia | SVO |
| Swahili | Niger-Congo | Tanzania | SVO |
| Yoruba | Niger-Congo | Nigeria, Benin | SVO |
| English | Indo-European | Various | SVO |

Table 10: Linguistic characteristics of selected languages.

## C  Unique Concepts in Different Languages

Table 11 displays the number of unique concepts and some examples in each language after basic deduplication (lemmatization and casing).

| Lang. | Unique Concepts | Examples |
|---|---|---|
| hi | 494 | samosa<br>seasonal rain<br>sweet gulkand |
| id | 742 | smell of durian<br>young rambutan<br>shinchan |
| jv | 303 | elephant riding rickshaw<br>sugar cane<br>tripe skin |
| kn | 444 | dosa<br>ayurveda<br>buddha's smile |
| su | 365 | sticky rice<br>papaya tree<br>lotus flower in water |
| sw | 481 | baobab<br>king solomon<br>clove ointment |
| yo | 333 | president buhari<br>rock of olumu<br>aníkúlápó movie |
| en | 954 | thanksgiving buffet<br>washington post reporter<br>renaissance artist |

Table 11: Number and examples of unique object concepts expressed in each language (translated to EN). Unique concepts here are those not shared by any other language in the dataset.

## D  Jaccard Similarity between Concepts

Table 12 contains Jaccard similarities for sets of concepts found in each language. Language pairs with the highest similarity (row-wise) are bolded.

## E  Training Details

A hyperparameter grid search was conducted over values: epochs $\in \{10, 20, 30\}$, lr $\in \{2 \times 10^{-4}, 5 \times 10^{-4}, 2 \times 10^{-5}, 5 \times 10^{-5}, 2 \times 10^{-6}, 5 \times 10^{-6}\}$, and batch size $\in \{32, 64\}$.

XLM-R$_{\text{large}}$ was trained for 20 epochs with a learning rate of $5 \times 10^{-6}$ and a batch size of 32. XLM-R$_{\text{large}}$ was trained for 30 epochs with a learning rate of $2 \times 10{-5}$ and a batch size of 64. mBERT$_{\text{base}}$ was trained for 30 epochs with a learning rate of $5 \times 10^{-5}$ and a batch size of 64. An A6000 GPU was used for each model. Each training run takes on the order of a few minutes.

Most seeds lead to a near-random performance on the English dev set, while a small minority of seeds lead to non-random performance. We took the top 5 seeds from 1-100 found in terms of English dev set performance in order to avoid including results from degenerate seeds.

We did not experiment with trying to optimize the hyperparameters for the experiments in Section 5.2.2 but rather used the same ones found previously. This may account for some settings leading to lower performance.

## F  Few-shot Full Results

Table 13 outlines the effect of adding $k \in \{2, ..., 50\}$ examples in each target language.

## G  Four-Quadrant Examples

### G.0.1  Easy

- **नदी का पानी क्रिस्टल की तरह साफ है**/the water of the river is as clear as crystal

- **Ia berjalan layaknya siput**/he walks like a snail

- **Inú yàrá ìdánwò nàá palọ́lọ́ bí i itẹ́ òkú**/inside the exam room was a dead silence

- **Vijana ndio taifa la kesho**/youth is the nation of tomorrow

|      | hi     | id     | jv     | kn     | su     | sw     | yo     | en     |
|------|--------|--------|--------|--------|--------|--------|--------|--------|
| hi   | -      | 0.0477 | 0.0541 | **0.0945** | 0.0534 | 0.0904 | 0.0509 | 0.0631 |
| id   | 0.0477 | -      | **0.0588** | 0.0431 | 0.0405 | 0.0544 | 0.0352 | 0.0425 |
| jv   | 0.0541 | 0.0588 | -      | 0.0619 | 0.067  | **0.0724** | 0.0449 | 0.0377 |
| kn   | **0.0945** | 0.0431 | 0.0619 | -      | 0.0464 | 0.0842 | 0.0594 | 0.0586 |
| su   | 0.0534 | 0.0405 | **0.067** | 0.0464 | -      | 0.0563 | 0.0444 | 0.0312 |
| sw   | **0.0904** | 0.0544 | 0.0724 | 0.0842 | 0.0563 | -      | 0.0671 | 0.0693 |
| yo   | 0.0509 | 0.0352 | 0.0449 | 0.0594 | 0.0444 | **0.0671** | -      | 0.0311 |
| en   | 0.0631 | 0.0425 | 0.0377 | 0.0586 | 0.0312 | **0.0693** | 0.0311 | -      |

Table 12: Jaccard similarities between object sets for each language. The language that is most similar is bolded for each row.

|  | Lang. | $k=2$ Score | $\Delta$ | $k=10$ Score | $\Delta$ | $k=20$ Score | $\Delta$ | $k=30$ Score | $\Delta$ | $k=40$ Score | $\Delta$ | $k=50$ Score | $\Delta$ |
|---|---|---|---|---|---|---|---|---|---|---|---|---|---|
| XLM-R$_{large}$ | hi | 67.47 | -0.11 | 67.47 | -0.11 | 67.29 | -0.29 | **67.72** | 0.14 | 67.67 | 0.09 | 67.58 | 0 |
|  | id | 78.01 | -0.08 | 78.04 | -0.05 | 78.22 | 0.13 | 77.91 | -0.18 | 78.04 | -0.05 | **78.56** | 0.47 |
|  | jv | 60.77 | -0.16 | **61.14** | 0.2 | 60.36 | -0.58 | 60.78 | -0.16 | 61.08 | 0.14 | 60.76 | -0.17 |
|  | kn | 58.09 | 0.01 | 58.17 | 0.09 | 59.34 | 1.26 | 59.38 | 1.3 | 60.39 | 2.31 | **61.91** | 3.83 |
|  | su | 60.47 | 0.07 | 60.55 | 0.15 | **61.36** | 0.96 | 60.22 | -0.18 | 60.35 | -0.05 | 61.28 | 0.88 |
|  | sw | 58.23 | 0.07 | 58.16 | 0 | 58.49 | 0.33 | 58.88 | 0.72 | 58.92 | 0.76 | **59.00** | 0.84 |
|  | yo | - | - | - | - | - | - | - | - | - | - | - | - |
| XLM-R$_{base}$ | hi | 62.47 | -0.01 | **62.51** | 0.03 | 62.27 | -0.21 | 62.45 | -0.03 | 62.06 | -0.42 | 61.89 | -0.59 |
|  | id | **69.23** | 0.35 | 69.07 | 0.19 | 69.16 | 0.28 | 69.20 | 0.32 | 68.66 | -0.22 | 69.14 | 0.26 |
|  | jv | 54.09 | 0.43 | 54.31 | 0.64 | 54.04 | 0.37 | 54.53 | 0.86 | 53.92 | 0.25 | **54.60** | 0.93 |
|  | kn | 54.62 | -0.04 | 54.55 | -0.12 | 54.56 | -0.11 | 54.53 | -0.14 | **55.05** | 0.38 | 54.44 | -0.22 |
|  | su | 51.95 | -0.46 | 51.90 | -0.51 | 51.72 | -0.69 | 51.37 | -1.03 | 51.27 | -1.14 | 50.48 | -1.93 |
|  | sw | 52.78 | 0.05 | 52.76 | 0.03 | 53.00 | 0.27 | 53.04 | 0.31 | 53.50 | 0.76 | **53.83** | 1.10 |
|  | yo | - | - | - | - | - | - | - | - | - | - | - | - |
|  |  | $k=2$ |  | $k=4$ |  | $k=6$ |  | $k=8$ |  | $k=10$ |  | $k=50$ |  |
| mBERT$_{base}$ | hi | 51.43 | 0.11 | 51.41 | 0.09 | **53.42** | 2.10 | 51.50 | 0.18 | 51.47 | 0.15 | 50.93 | -0.39 |
|  | id | 56.59 | 0.02 | 56.57 | 0.01 | 56.58 | 0.01 | **56.62** | 0.05 | 56.59 | 0.03 | 56.50 | -0.07 |
|  | jv | **55.13** | 0.07 | 55.03 | -0.03 | 54.93 | -0.13 | 55.00 | -0.06 | 54.86 | 0.20 | 54.64 | -0.42 |
|  | kn | 52.70 | 0.07 | 52.67 | 0.04 | **52.70** | 0.07 | 52.66 | 0.03 | 52.67 | 0.04 | 52.42 | -0.20 |
|  | su | 52.83 | -0.04 | 52.91 | 0.04 | 52.79 | -0.07 | 52.54 | -0.32 | 52.68 | -0.19 | **55.20** | 2.33 |
|  | sw | 52.12 | 0 | 52.13 | 0.01 | 52.14 | 0.02 | **52.20** | 0.08 | 52.15 | 0.03 | 51.76 | -0.36 |
|  | yo | 50.52 | -0.02 | 50.50 | -0.10 | 50.42 | -0.19 | 50.31 | -0.21 | 50.37 | -0.15 | 50.35 | -0.17 |

Table 13: Effect of adding additional examples in the target language to English training data. The highest improvement is bolded for each language.

- **Dia menjalani hidup bak singa di kebun binatang**/he lives life like a lion in the zoo

### G.0.2 Challenge - linguistic

- **Àgbẹ̀ náà pa gbogbo ọmọ tí igi nàá bí lánàá**/the farmer killed all the children that the tree gave birth to yesterday

- **Penzi lao ni kama moto wa kibatari kwenye upepo**/their love is like fire in the wind

- **Kadang jelema teh bisa ipis kulit bengeut**/sometimes people can have thin skin

- **Si eta kuliah siga nu teu kantos bobo**/that college guy looks like he never sleeps

- **ಅವರು ನೇಡೆದ್ದ ನೇರು ಸಮುದ್ರದ ನೇರೊನಂತೆ ಉಪ್ಪಾಗಿತ್ತು**/the water they gave was as salty as sea water

### G.0.3 Challenge - translation

- **hirup teh kudu boga kaditu kadieu**/life must have here and there

- **लड़की का व्यक्तित्व गुलाब जामुन की तरह मीठा था**/the girl's personality was as sweet as Gulab Jamun

- **Ìṣọ̀lá má ń tún ilé rẹ̀ ṣe ní gbogbo nìgbà**/honor does not repair his house all the time

- **Nek gawe wedang kopi Painem kaya disoki suruh**/if you make a Painem coffee drink, it's like being told

- **Bapak tirine sifate kaya Gatot Kaca**/his stepfather is like Gatot Kaca

### G.0.4 Hard

- कालिदास भारत के शेख्चिली हैं।/Kalidas is Shekhchili of India

- उसके मन का मैल मिटी की तरह छलनी से निकल गया।/The filth of his mind was removed from the sieve like soil

- **Wajahku dan adikku ibarat pinang di belah dua**/My face and my sister are like areca nuts split in half.

- **Hari ini cuacanya seperti berada di di puncak gunung Bromo**/Today the weather is like being at the top of Mount Bromo

- **Doni karo Yanti pancen kaya Rahwana Sinta ing pewayangan**/Doni and Yanti are really like Ravana Sinta in a puppet show

## H Accuracy on Annotated Commonsense Categories

Table 14 shows the accuracy on commonsense categories across all languages for XLM-R$_{\text{large}}$. Note that Yoruba is not included due to XLM-R$_{\text{large}}$ not being trained on this language.

| Language | Category | Acc. |
|---|---|---|
| hi | obj | 67.50 |
|  | vis | 67.48 |
|  | soc | 67.86 |
|  | cul | 70.65 |
| id | obj | 76.60 |
|  | vis | 76.56 |
|  | soc | 82.71 |
|  | cul | 77.11 |
| jv | obj | 65.02 |
|  | vis | 58.89 |
|  | soc | 64.48 |
|  | cul | 50.82 |
| kn* | obj | 57.14 |
|  | vis | 36.36 |
|  | soc | 55.56 |
|  | cul | 77.78 |
| su | obj | 57.07 |
|  | vis | 56.86 |
|  | soc | 67.50 |
|  | cul | 61.11 |
| sw | obj | 58.06 |
|  | vis | 61.99 |
|  | soc | 56.50 |
|  | cul | 52.46 |
| yo | obj | 48.15 |
|  | vis | 52.38 |
|  | soc | 49.58 |
|  | cul | 47.37 |

Table 14: Performance of XLM-R$_{\text{large}}$ on commonsense categories indicated by annotators.[12]